\documentclass[letterpaper, 10 pt, conference]{ieeeconf}  

\IEEEoverridecommandlockouts                              

\usepackage{graphicx}
\usepackage{float}
\usepackage{amsmath}
\usepackage[section]{placeins}
\usepackage{lipsum} 
\usepackage{environ} 
\usepackage{amsmath}
\usepackage{tabularx}
\usepackage{adjustbox}

\title{Improving Visual Place Recognition Performance by Maximising Complementarity}

\author{Maria Waheed$^{1}$,  Michael Milford$^{2}$,  Klaus McDonald-Maier$^{1}$ and Shoaib Ehsan$^{1}$
\thanks{*This work was supported by the UK Engineering and Physical Sciences Research Council through Grants EP/R02572X/1, EP/P017487/1 and in part by the RICE project funded by the National Centre for Nuclear Robotics Flexible Partnership
Fund. \textit{(Corresponding author: Shoaib Ehsan)}}
\thanks{$^{1}$M. Waheed, K. McDonald-Maier and S. Ehsan are with the School of Computer Science and Electronic Engineering, University of Essex, Colchester CO4 3SQ, United Kingdom
        {\tt\small (e-mail: mw20987@essex.ac.uk; kdm@essex.ac.uk; sehsan@essex.ac.uk)}}%
\thanks{$^{2}$M. Milford is with the School of Electrical Engineering and Computer
Science, Queensland University of Technology, Brisbane, QLD 4000, Australia
        {\tt\small (e-mail: michael.milford@qut.edu.au)}}%
}

\begin{document}

\maketitle
\thispagestyle{empty}
\pagestyle{empty}

\begin{abstract}

Visual place recognition (VPR) is the problem of recognising a previously visited location using visual information. Many attempts to improve the performance of VPR methods have been made in the literature. One approach that has received attention recently is the multi-process fusion where different VPR methods run in parallel and their outputs are combined in an effort to achieve better performance. The multi-process fusion, however, does not have a well-defined criterion for selecting and combining different VPR methods from a wide range of available options. To the best of our knowledge, this paper investigates the complementarity of state-of-the-art VPR methods systematically for the first time and identifies those combinations which can result in better performance. The paper presents a well-defined framework which acts as a sanity check to find the complementarity between two techniques by utilising a McNemar's test-like approach. The framework allows estimation of upper and lower complementarity bounds for the VPR techniques to be combined, along with an estimate of maximum VPR performance that may be achieved. Based on this framework, results are presented for eight state-of-the-art VPR methods on ten widely-used VPR datasets showing the potential of different combinations of techniques for achieving better performance.

\end{abstract}

\section{Introduction}

Visual place recognition is a fundamental yet challenging task in the field of mobile robotics [1]. It may be defined as the ability of a robot to recognize a previously visited location. Viewpoint changes [2], [3], seasonal variations [4], [5], presence of dynamic objects [6], [7]  and illumination changes [8], [9] encountered in real world scenarios make this apparently simple task non-trivial [4], [10]. Several techniques have been presented to solve this problem (such as [11]-[14]), however, every VPR method has its own pros and cons [15]-[18], and there is no universal technique that may be used in all conditions and scenarios . \\

\begin{figure}[htb!]
    \centering
    \scalebox{0.5}
    
    \includegraphics[width=6.5cm,height=5cm]{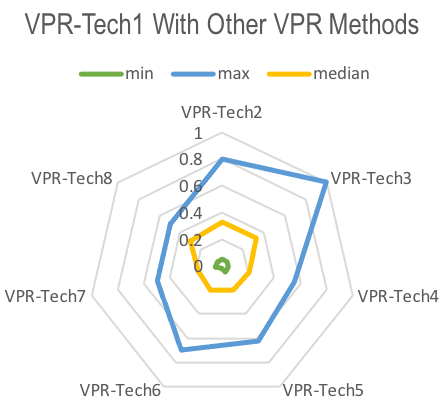}
    \caption{Sample output of the proposed complementarity framework. Here, VPR-Tech1 is the primary VPR technique which may be combined with other available secondary VPR methods (VPR-Tech2, VPR-Tech3 etc). The green line (\textbf{min}) shows the lower complementarity bound of VPR-Tech1 with other methods; the blue line (\textbf{max}) depicts the maximum complementarity bound; the yellow line (\textbf{median}) shows the median complementarity bound.}
    \label{fig:my_label}
\end{figure}

Recently, a new approach named multi-process fusion has been introduced that combines several image processing methods and negates the requirement of multiple sensors to improve VPR performance [19], [20]. The concept comes from the empirical data which suggests that some VPR methods are more suitable for certain types of environments and scenarios than others [10]. Hence, utilising multiple VPR techniques simultaneously may compensate for each other’s weaknesses. Although the systems presented in [19], [20] exhibit promising results, they do not provide a well-defined criterion for selection of VPR techniques based on complementarity out of the available options. Supposing that the fused VPR methods will complement each other in all cases is not a valid assumption and may have detrimental effect on performance and computation. For example, if the VPR techniques that are combined are redundant, they will not achieve higher performance and will only add to the computational cost which may not be suitable for resource-constrained systems. Hence, complementarity information is vital and can enable a multi-process fusion based system to make an informed decision regarding selection of VPR techniques from available options.\\

To the best of our knowledge, complementarity of VPR methods has not been studied systematically so far. Through this paper, we attempt to bridge this gap and intend to design a framework that can be used as a sanity check for the selection of complementary pairs of VPR techniques for multi-process fusion systems. Our proposed framework is based on a McNemar's test-like approach [21], [22] that categorizes each VPR outcome from a technique as either success or failure (considering ground truth information). The framework allows estimation of upper and lower complementarity bounds for the VPR techniques to be combined, along with an estimate of maximum VPR performance that may be achieved. This framework is then employed for eight state-of-the-art VPR methods to identify highly complementary pairs on widely used VPR data sets. \\
The rest of this paper is organized as follows. Section II provides an overview of related work. Section III presents the framework for computing complementarity between VPR techniques, and for estimating upper and lower complementarity bounds along with an assessment of maximum achievable VPR performance. Section IV describes the experimental setup. The results based on the proposed framework are presented in Section V. Finally, conclusions are given in section VI.

\section{Related Work}
This section provides an overview of the related work in the domain of visual place recognition. The methods used for VPR may be divided into three categories: handcrafted feature descriptor-based techniques, deep-learning-based methods, and Region-of-interest-based approaches. All these categories have their own strengths and weaknesses that influence the selection of any methods from among them. Some state-of-the-art handcrafted feature descriptors used for VPR are Scale Invariant Feature Transform (SIFT) [23], Speeded-Up Robust Features (SURF) [24], and GIST [25]. Convolutional Neural Networks (CNNs) have turned out to be revolutionary in the field of VPR and provide significant improvement in performance [26] even under extreme environmental variations. Some of the widely used techniques include NetVLAD [11], AMOSNet [12], and HybridNet [12]. Region-of-interest-based VPR techniques make use of the static and definite regions of images to perform place recognition, such as Regions of Maximum Activated Convolutions (R-MAC) [27]. \\
Fusing multiple sensors to improve place recognition performance has been the focus of several research works [28]-[30]. Although multi-sensor approaches help boost performance, they do carry certain disadvantages, such as expensive and bulky sensors, and potential significant increase in computation. To overcome these shortcomings, the concept of fusing multiple VPR techniques has gained popularity. The authors of [31] combined multiple image processing methods into a merged feature vector using a convex optimization approach to decide the best match from the sequence of images generated. The effort did generate some promising results over multiple datasets but had limited overall performance due to the absence of sequential information. Similarly, a multi-process fusion system is introduced in [19] which combines multiple VPR methods using a Hidden Markov Model (HMM) to identify the optimal estimated location over a sequence of images. 
\begin{figure}[!htb]
    \centering
    \vspace*{0.1in}
    \includegraphics[width=1\columnwidth]{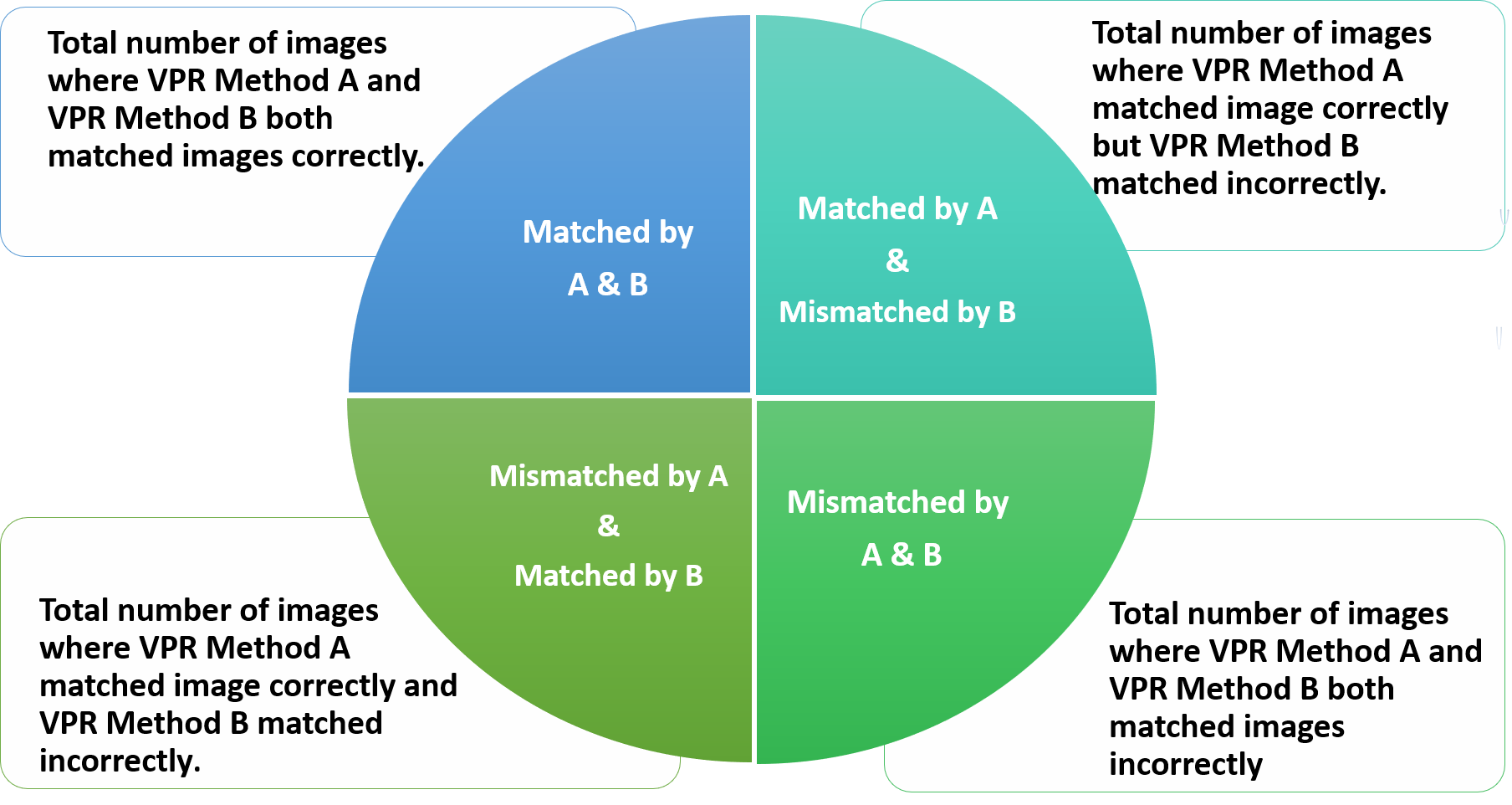}
    \caption{Possible outcomes of pairwise analysis of VPR methods on a case-by-case basis over the same dataset.}
    \label{fig:my_label}
\end{figure}
The authors of [20] have presented a three-tier hierarchical multi-process fusion system which is customizable and may be extended to any arbitrary number of tiers. A different place recognition 
method is used in each tier to compare the query image with the provided sequence of images.

\section{Proposed Framework}
This section presents the framework for computing complementarity, for establishing the upper and lower complementarity bounds, and for estimating the maximum achievable VPR performance by a muti-process fusion system. This framework may be employed on an arbitrary number of VPR methods to determine the optimal pairing from among 
the pool of techniques available.It may also be utilised as a sanity check on whether the VPR techniques that a multi-process fusion system has assembled for integration are even viable. The framework employs a McNemar’s test like approach to perform a case-by-case analysis of each VPR technique to compute the complementarity of the given technique with other available methods.

Precision-recall curves, F-scores and accuracy percentage [9] are usually utilised as performance metrics for VPR methods. Although viable for some applications / scenarios, these performance metrics do not provide the specific information that tells where exactly does a VPR method succeeds or fails, and do not show the whole picture. For example, two VPR methods compared over a dataset of 100 images using these performance metrics may appear to have same performance if they both are able to match 70 images (out of 100). However, it is highly likely that the set of 70 images successfully matched by the first VPR method is not the same set that is also correctly matched by the second VPR technique. We believe that this neglected piece of information is critical for determining complementarity of different VPR methods, and is vital knowledge to have specifically when dealing with multi-process fusion systems.

McNemar's test is a form of chi-squared test with one degree of freedom that evaluates the performance of two algorithms based on their outcomes on a case-by-case basis over the same dataset. For utilizing McNemar’s test, a criterion is needed to determine whether a test case results in success or failure. Our proposed framework is loosely inspired by the McNemar's test as we do pairwise analysis of VPR methods on a case-by-case basis over the same dataset. The two VPR methods in question would produce results in the form of correct or incorrect matches verified using ground truth. This data may then be divided into four possible cases as shown in Fig. 2: first being the number of images where both algorithms are able to match the images correctly, second where the first algorithm matched correctly while the second produced an incorrect match, then vice versa and finally where both algorithms failed and produced incorrect matches. For computing complementarity, our prime focus remains on case two and three as these hold the number of images where the two algorithms perform differently and can help boost each other’s performance.

\textbf{Computing complementarity.} Let \textit{A} be our primary VPR technique. Let \textit{B} be a VPR method that may be combined with \textit{A} in a multi-process fusion system to enhance VPR performance over an image dataset \textit{D}. VPR performance is defined as the ratio of number of images of \textit{D} that are correctly matched (verified by groundtruth) to the total number of images of \textit{D}. The complementarity is calculated by the following equation:  

\begingroup
\small
\begin{equation}
CBA = \frac{T} {M}\\
\end{equation}
\endgroup

Where \textit{CBA} is the complementarity of \textit{B} with \textit{A}; \textit{T} is the number of images of \textit{D} which are incorrectly matched by \textit{A} but correctly matched by \textit{B} when the two methods are run; \textit{M} is the number of images of \textit{D} that are incorrectly matched by \textit{A} when run. A large value of \textit{CBA} implies that \textit{B} complements \textit{A} well on dataset \textit{D} and will result in potential increase in VPR performance. On the other hand, a small value of \textit{CBA} means that \textit{B} does not complement \textit{A} well. In other words, \textit{A} and \textit{B} are redundant and combining \textit{A} with \textit{B} will increase computational cost without any substantial increase in VPR performance. 

\textbf{Establishing complementarity bounds.} It is interesting to further explore the upper and lower extremities of complementarity of \textit{B} with \textit{A}. Let \textit{K} be the set of \textit{n} individual datasets on which \textit{A} and \textit{B} are run. 

\begin{equation}
    \centering
    K = \{D_1, D_2, D_3,….D_n\}
\end{equation}
			
Let \textit{J} be the set of complementarity scores (\textit{B} with \textit{A}) computed over \textit{n} dataset in \textit{K}.
    \begin{equation}
        \centering
        J = \{CBA_1, CBA_2, CBA_3…CBA_n\}
    \end{equation}

The upper complementarity bound is then established as

\begin{equation}
    \centering
    U = max \{CBA_1, CBA_2, CBA_3…CBA_n\}
\end{equation}

The lower complementarity bound is estimated as

        \begin{equation}
           \centering
	    	L = min \{CBA_1, CBA_2, CBA_3…CBA_n\}
		\end{equation}

The median of complementarity of \textit{B} with \textit{A} is computed as 
       \begin{equation}
           \centering
           Q = median \{CBA_1, CBA_2, CBA_3…CBA_n\}
       \end{equation} 

\textbf{Estimating maximum achievable performance.} It is beneficial to estimate the maximum achievable VPR performance of a multi-process fusion system over a dataset at an early stage. This is estimated as follows:

\begingroup
\small
\begin{equation}
\small
        MAPE = \frac{(T + W + X )}{Y}
\end{equation}
\endgroup

Where \textit{MAPE} is the maximum achievable VPR performance estimate for the multi-process fusion system over a dataset \textit{D}; \textit{T} is the number of images of \textit{D} which are incorrectly matched by \textit{A} but correctly matched by \textit{B} when the two methods are run; \textit{W} is the number of images of \textit{D} which are correctly matched by \textit{A} but incorrectly matched by \textit{B} when the two methods are run; ; \textit{X} is the number of images of \textit{D} which are correctly matched by both \textit{A} and \textit{B} when the two methods are run; \textit{Y} is the total number of images of \textit{D}.

\section{Experimental Setup}

This section provides details of the experimental setup used for obtaining results by utilising the proposed framework. Table 1 lists the widely used VPR datasets [31] that are used for our experiments, namely GardensPoint, 24/7 Query [33], Essex3in1 [34], SPEDTest, Cross-Seasons [35], Synthia [36], Corridor, 17-Places, Living room, and Nordland [37]. The implementation details of the eight state-of-the-art VPR techniques that are utilised in the experiments are given below.  

\textbf{AlexNet:} The use of AlexNet for VPR was studied by [40], who suggested that \textit{conv3} is the most robust to conditional variations. Gaussian random projections are used to encode the activation-maps from conv3 into feature descriptors. Our implementation of AlexNet is similar to the one employed by [41].

\textbf{NetVLAD:} The original implementation of NetVLAD was in MATLAB, as released by [11]. The Python part of this code was open-sourced by [38]. The model selected for evaluation is VGG-16, which has been trained in an end-to-end manner on Pittsburgh 30K dataset [11] with a dictionary size of 64 while performing whitening on the final descriptors.

\textbf{AMOSNet:} This technique was proposed by [12], where a CNN was trained from scratch on the SPED dataset. The authors presented results from different convolutional layers by implementing spatial pyramidal pooling on the respective layers. While the original implementation is not fully open-sourced, the trained model weights are shared by authors.

\textbf{HybridNet:} While AMOSNet was trained from scratch, [12] took inspiration from transfer learning for HybridNet and re-trained the weights initialised from Top-5 convolutional layers of CaffeNet [39] on SPED dataset. We have implemented HybridNet using 'conv5' of the shared HybridNet model.

\begin{figure*}[!htb]
    \vspace*{0.1in}
    \includegraphics[width=2\columnwidth]{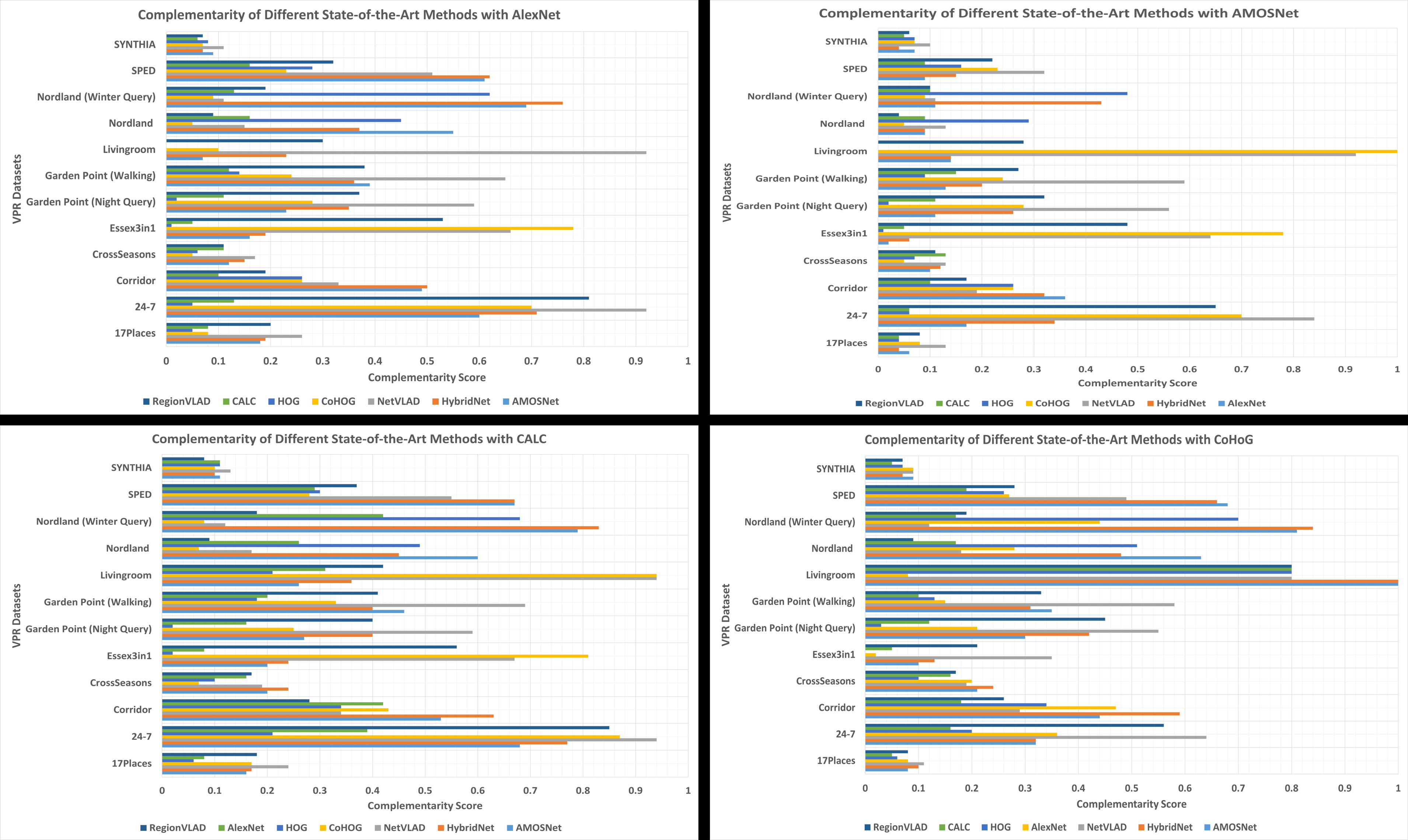}
    \caption{Complementarity of state-of-the-art VPR methods with: AlexNet (top left); AMOSNet (top right); CALC (bottom left); CoHoG (bottom right).}
    \label{figurelabel}
\end{figure*}

\textbf{RegionVLAD:} This technique is introduced and open-sourced by [14]. We have used AlexNet (trained on Places365 dataset) as the underlying CNN. The total number of regions of interest is set to 400, and we have used ‘conv3’ for feature extraction. The dictionary size is set to 256 visual words for VLAD retrieval. Cosine similarity is subsequently used for matching descriptors of query and reference images

\textbf{CALC:} The use of convolutional auto-encoders for VPR was proposed by [13], where an auto-encoder network was trained in an unsupervised manner to re-create similar HOG descriptors for viewpoint variant (cropped) images of the same place. We use model parameters from 100,000 training iteration. Cosine-matching is used for descriptor comparison.

\textbf{HoG:} Histogram-of-oriented-gradients (HoG) is one of the most widely used handcrafted feature descriptor, which actually performs very well for VPR compared to other handcrafted feature descriptors. We use a cell size of $16 \times 16$ and a block size of $32 \times 32$ for an image-size of $512 \times 512$ for our implementation. The total number of histogram bins are set equal to 9. We use cosine-matching between HOG-descriptors of various images to find the best place match.

\textbf{CoHoG:} It is a recently proposed handcrafted feature descriptor-based technique, which uses image-entropy for region-of-interest extraction. The regions are subsequently described by dedicated HoG descriptors and these regional descriptors are convolutionally matched to achieve lateral viewpoint-invariance. It is an opensource technique and we have used an image size of $512 \times 512$, cell size of $16 \times 16$, bin-size of 8 and an entropy-threshold (ET) of 0.4. CoHoG also uses cosine-matching for descriptor comparison.

\section{Results and Discussion}

\begin{figure*}[!htb]
    \vspace*{0.1in}
    \includegraphics[width=2\columnwidth]{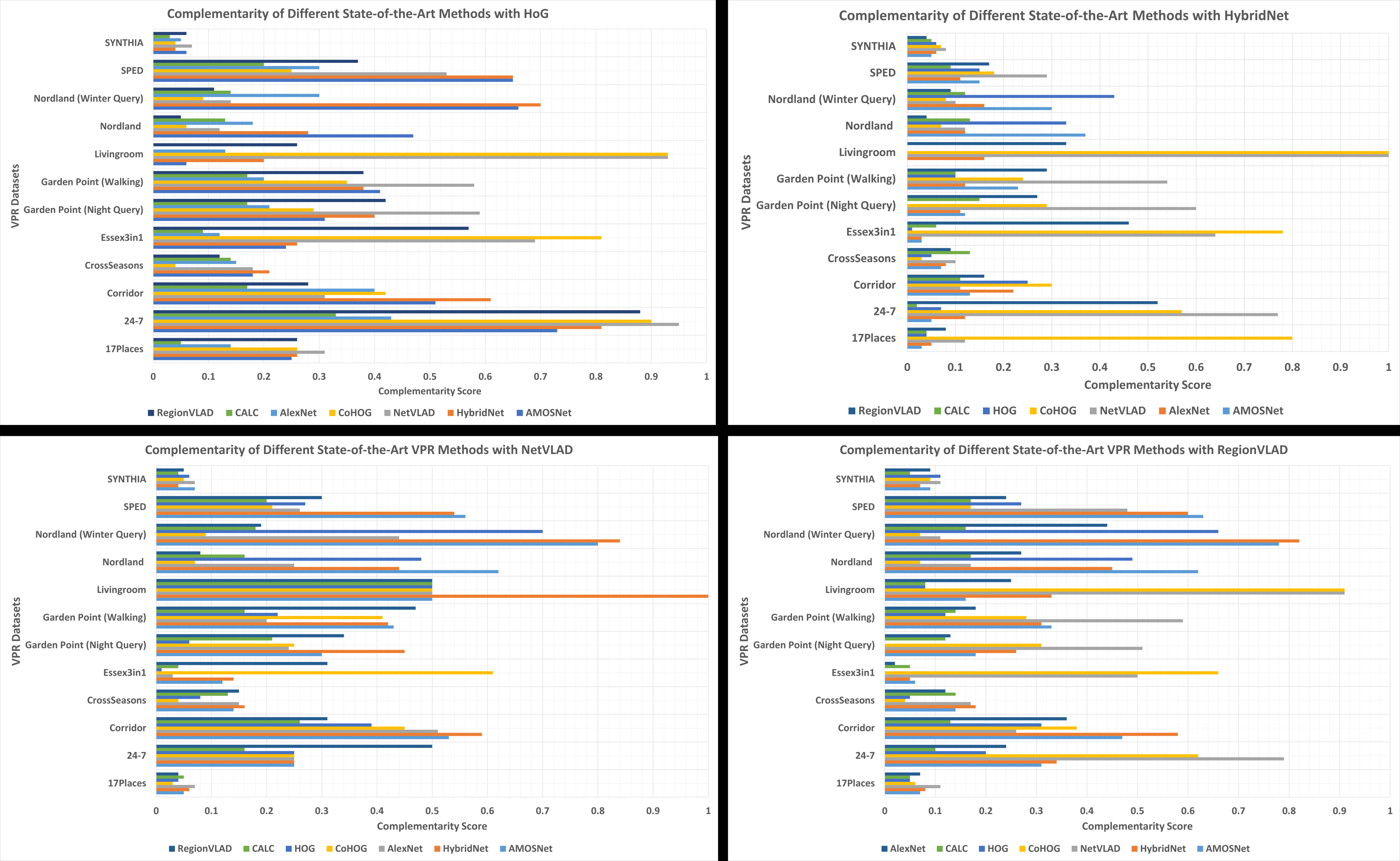}
    \caption{Complementarity of state-of-the-art VPR methods with: HoG (top left); HybridNet (top right); NetVLAD (bottom left); RegionVLAD (bottom right).}
    \label{figurelabel}
\end{figure*}

\begin{table*}[]
\centering
\caption{VPR-Bench Datasets[32]}
\begin{tabular}{|c|c|c|c|c|c|}
\hline
\textbf{Dataset} & \textbf{Environment} & \textbf{Query   Images} & \textbf{Ref Images} & \textbf{Viewpoint-Variation} & \textbf{Conditional-Variation} \\ \hline
GardensPoint     & University Campus    & 200                     & 200                 & Lateral                      & Day-Night                      \\ \hline
24/7 Query       & Outdoor              & 375                     & 750                 & 6-DOF                        & Day-Night                      \\ \hline
ESSEX3IN1        & University Campus    & 210                     & 210                 & 6-DOF                        & Illumination                   \\ \hline
SPEDTest         & Outdoor              & 607                     & 607                 & None                         & Seasonal and Weather           \\ \hline
Cross-Seasons    & City-Like            & 191                     & 191                 & Lateral                      & Dawn-Dusk                      \\ \hline
Synthia          & City-like(Synthetic) & 947                     & 947                 & Lateral                      & Seasonal                       \\ \hline
Nordland         & Train Journey        & 1622                    & 1622                & None                         & Seasonal                       \\ \hline
Corridor         & Indoor               & 111                     & 111                 & Lateral                      & None                           \\ \hline
17-Places        & Indoor               & 406                     & 406                 & Lateral                      & Day-Night                      \\ \hline
Living-room      & Indoor               & 32                      & 32                  & Lateral                      & Day-Night                      \\ \hline
\end{tabular}
\end{table*}
%
%
\begin{table*}[htbp]
\vspace*{0.2in}
\caption{ Maximum achievable performance estimate for different combinations of state-of-the-art VPR methods on standard datasets in percentage}
\label{tab:caption}
\begin{adjustbox}{width=\textwidth}
\begin{tabular}{|l|l|l|l|l|l|l|l|l|l|l|l|l|}
\hline
\textbf{VPR Combinations} & \textbf{17Places} & \textbf{24-7} & \textbf{Corridor} & \textbf{CrossSeasons} & \textbf{Essex3in1} & \textbf{Garden Point} & \textbf{Livingroom} & \textbf{Nordland} & \textbf{SPED} & \textbf{SYNTHIA} \\ \hline
AlexNet + AMOSNet         & 43.1              & 87.2          & 73.8              & 32.9                  & 28.0               & 54.0                  & 62.5                & 83.6              & 81.3          & 32.2             \\ \hline
AlexNet + CALC            & 36.2              & 72.0          & 54.0              & 31.9                  & 19.0               & 34.0                  & 59.3                & 53.8              & 59.6          & 30.62            \\ \hline
AlexNet + CoHoG           & 44.0              & 95.7          & 71.17             & 28.7                  & 82.8               & 48.5                  & 96.8                & 50.8              & 63.4          & 32.7             \\ \hline
AlexNet + HoG             & 33.7              & 69.3          & 62.1              & 28.2                  & 15.7               & 36.0                  & 59.3                & 79.9              & 65.2          & 32.1             \\ \hline
AlexNet + HybridNet       & 43.3              & 90.6          & 74.7              & 35.0                  & 30.1               & 52.0                  & 68.7                & 87.4              & 81.7          & 31.3             \\ \hline
AlexNet + NetVLAD         & \textbf{48.2}     & 97.6          & 65.7              & \textbf{36.1}         & 70.9               & 65.5                  & 96.8                & 52.7              & 76.2          & \textbf{34.2}    \\ \hline
AlexNet + RegionVLAD      & 44.5              & 94.1          & 58.5              & 31.9                  & 60                 & 53.5                  & 71.8                & 57.1              & 67.2          & 31.2             \\ \hline
AMOSNet + CALC            & 41.8              & 85.6          & 63.0              & 35.0                  & 30.0               & 55.5                  & 56.2                & 83.4              & 81.3          & 30.4             \\ \hline
AMOSNet + CoHoG           & 44.0              & 95.4          & 69.3              & 29.3                  & 84.2               & 60.5                  & \textbf{100}        & 83.1              & 84.1          & 32.5             \\ \hline
AMOSNet + HoG             & 42.11             & 85.6          & 69.3              & 30.8                  & 27.1               & 52.5                  & 56.2                & 90.3              & 82.8          & 32.5             \\ \hline
AMOSNet + HybirdNet       & 42.11             & 89.8          & 72.0              & 34.5                  & 30.9               & 57.9                  & 62.5                & 89.5              & 82.5          & 29.8             \\ \hline
AMOSNet + NetVLAD         & 47.2              & 97.6          & 66.6              & 35.0                  & 73.8               & 75.5                  & 96.8                & 83.6              & \textbf{86.1} & 34.1             \\ \hline
AMOSNet + RegionVLAD      & 44.5              & 94.6          & 65.7              & 33.5                  & 61.9               & 62.0                  & 68.7                & 83.4              & 84.0          & 31.2             \\ \hline
CALC + CoHoG              & 42.3              & 94.3          & 54.9              & 25.1                  & 83.3               & 45.5                  & 96.8                & 26.6              & 59.1          & 29.6             \\ \hline
CALC +HoG                 & 34.4              & 63.4          & 47.7              & 27.2                  & 13.3               & 33.0                  & 53.1                & 75.0              & 60.1          & 30.6             \\ \hline
CALC + HybridNet          & 42.3              & 94.3          & 54.9              & 25.1                  & 83.3               & 45.5                  & 96.8                & 26.6              & 59.1          & 29.6             \\ \hline
CALC + NetVLAD            & 47.0              & 97.3          & 47.7              & 34.5                  & 71.4               & 63.5                  & 96.8                & 30.0              & 74.4          & 32.1             \\ \hline
CALC + RegionVLAD         & 43.3              & 93.0          & 43.2              & 32.9                  & 61.4               & 51.5                  & 65.6                & 35.1              & 64.4          & 28.4             \\ \hline
CoHoG + HoG               & 42.8              & 94.6          & 63.9              & 19.3                  & 82.3               & 47.5                  & 96.8                & 73.5              & 62.9          & 31.2             \\ \hline
CoHoG + HybridNet         & 45.0              & 95.4          & \textbf{77.4}     & 31.9                  & 84.7               & 58.5                  & \textbf{100}        & 86.2              & 83.1          & 31.4             \\ \hline
CoHoG + NetVLAD           & 45.8              & 97.6          & 61.2              & 27.2                  & \textbf{88.5}      & 74.5                  & 96.8                & 22.3              & 74.6          & 32.9             \\ \hline
CoHoG + RegionVLAD        & 44.0              & 97.0          & 59.4              & 25.6                  & 86.1               & 59.5                  & 96.8                & 28.4              & 64.0          & 31.5             \\ \hline
HoG + HybridNet           & 42.6              & 90.1          & 75.6              & 33.5                  & 29.5               & 50.5                  & 62.5                & \textbf{91.4}     & 82.5          & 31.3             \\ \hline
HoG + NetVLAD             & 45.8              & 97.6          & 61.2              & 27.2                  & \textbf{88.5}      & 74.5                  & 96.8                & 22.3              & 74.6          & 32.5             \\ \hline
HoG + RegionVLAD          & 43.3              & 93.8          & 54.9              & 26.1                  & 59.0               & 50.0                  & 65.6                & 74.3              & 68.6          & 32.5             \\ \hline
HybridNet + NetVLAD       & 45.0              & 94.9          & 72.9              & \textbf{36.1}         & 61.4               & 61.0                  & 75.0                & 86.4              & 83.0          & 29.7             \\ \hline
HybridNet + RegionVLAD    & 45.0              & 94.9          & 72.9              & \textbf{36.1}         & 61.4               & 61.0                  & 75.0                & 86.4              & 83.0          & 29.7             \\ \hline
NetVLAD + RegionVLAD      & 46.7              & \textbf{98.4} & 51.3              & 35.6                  & 79.5               & \textbf{77.0}         & 96.8                & 31.3              & 77.5          & 33.1             \\ \hline

\end{tabular}
\end{adjustbox}
\end{table*}

\begin{figure*}[!htb]
\centering
\vspace*{0.2in}
\includegraphics[width=2\columnwidth,height=13cm]{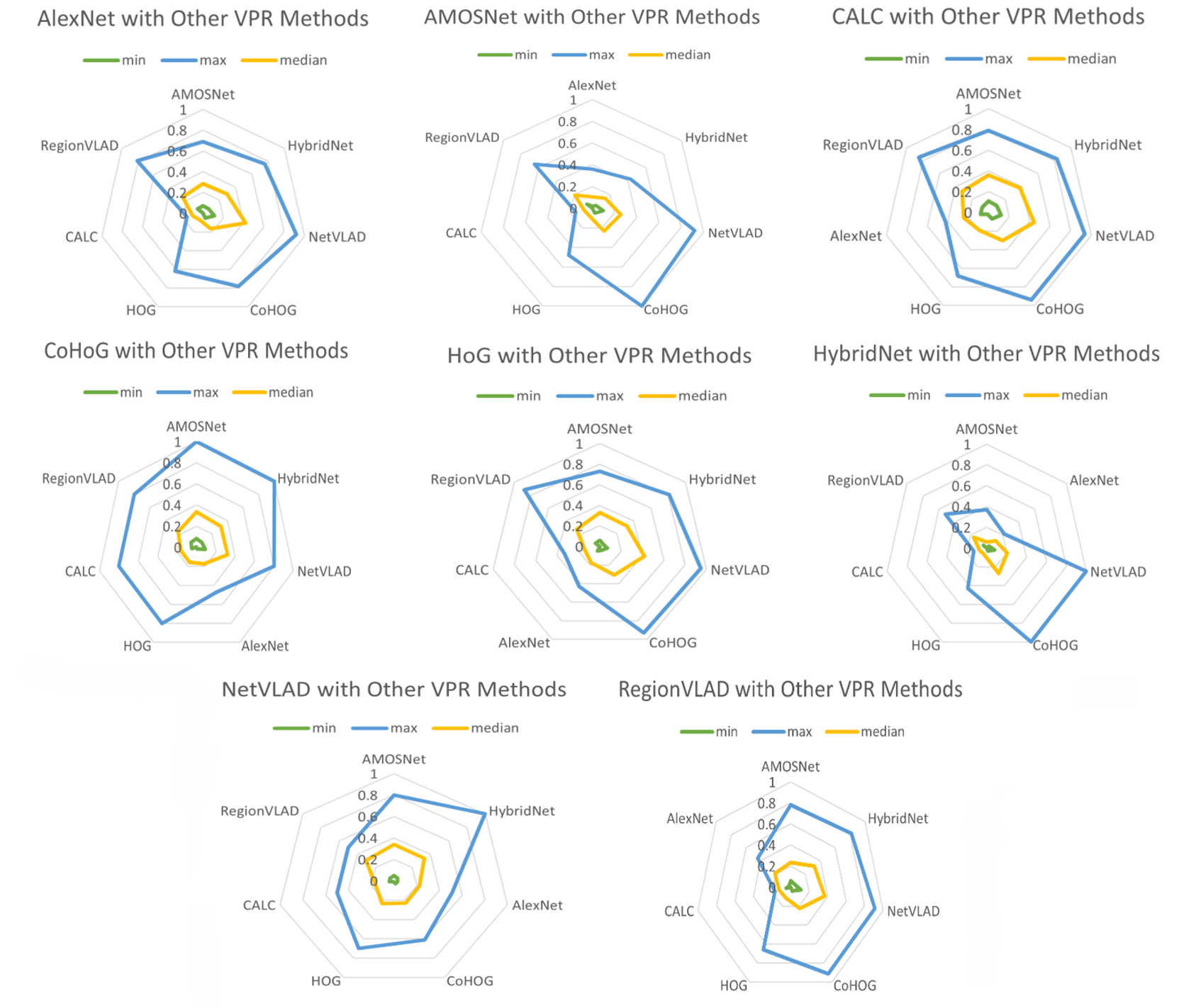}
\caption{Max (upper bound), Min (lower bound), and Median complementarity of state-of-the-art VPR methods with: AlexNet (top left); AMOSNet (top centre); CALC (top right); CoHoG (middle left); HoG (middle centre) HybridNet (middle right); NetVLAD (bottom left); RegionVLAD (bottom right).}
\label{figurelabel}
\end{figure*}

This section presents the results generated by utilising the proposed framework over a set of eight state-of-the-art VPR techniques on a variety of standard VPR data sets. Fig. 3 and Fig. 4 depict the complementarity scores of different VPR methods with each other on various standard data sets. Fig. 3 depicts how AlexNet has a high complementairty to NetVLAD, HybridNet and RegionVLAD while other methods have a significantly lower complementarity. This can be observed from 24-7, Essex3in1, GardenPoint and Livingroom datasets for NetVLAD (0.9, 0.65, 0.65 and 0.9 respectively). While the combination with HybridNet performs well on 24-7, Corridor, Nordland and SPED data sets (0.7, 0.5, 0.75 and 0.6 respectively) and the combination with RegionVLAD has the highest scores on 24-7 and Essex3in1 dataset. The remaining options are either not as complementary to AlexNet or perform well for only one data set but not the others. 
Except CoHoG and NetVLAD all other methods considered do not complement AMOSNet well. This can be noticed from the high complementarity scores of AMOSNet with CoHoG on 24/7 and Essex3in1 and Livingroom datasets (0.7, 0.8 ,1 respectively). Similarly, the combination of AMOSNet with NetVLAD achieves complementarity scores of 0.85, 0.65, 0.9 for the same three datasets respectively. Any one of the two techniques (CoHoG and NetVLAD) would be a viable fit for combining with AMOSNet in a system, while the least fit technique to be fused with AMOSNet on the basis of complementarity appears to be CALC which consistently scores low on all the datasets. 
Considering CALC as the primary VPR technique, it is evident from Fig. 3 that there are several suitable options with high complementarity scores over several data sets that are available. For example, CoHoG complements CALC well as shown by high scores on 24/7, Essex3in1, Livingroom and SPED datasets with exceptions present for Cross-Seasons and Nordland datasets. The second-best option for combining with CALC appears to be HybridNet which achieves high complementarity scores for several datasets including 24-7, Corridor, Nordland and SPED. Another promising technique to combine with CALC as the primary technique is NetVLAD as it performs as well as CoHoG on 24-7, Essex3in1, GardenPoint and Livingroom datasets (0.9, 0.6, 0.7, 0.9 respectively). The least complementary techniques with CALC appear to be AlexNet and RegionVLAD as evident from low scores over all datasets. 
When utilising CoHoG as the primary VPR technique, fusing HybridNet, AMOSNet and NetVLAD seem to be suitable options (see Fig. 3) and the least favorable option appears to be CALC. 

Fig. 4 displays the complementarity of HoG as the primary technique and presents several viable options including CoHoG, NetVLAD, HybridNet and AMOSNet. On the other hand, HybridNet as the primary technique has some substantially varying results with high complementarity scores achieved by CoHoG and NetVLAD reaching close to 0.8 for 24/7, Essex3in1 and Livingroom datasets. However, combinations with AMOSNet and CALC mostly have low scores remaining below 0.4 and 0.1 for all datasets. Combinations formed with NetVLAD (as the primary VPR technique), the complementarity scores indicate that different combinations work well for different datasets although HybridNet seems to outshine other combinations on Livingroom and Nordland datasets along with substantial scores for other datasets as well. The consistently low complementarity scores belong to RegionVLAD and CALC, thus making them the least suitable options with NetVLAD. Finally, RegionVLAD complements well with CoHoG, NetVLAD and HybridNet while having the lowest complementarity with CALC and AlexNet. It may be concluded from Fig. 3 and Fig. 4 that the most suitable VPR technique to form viable combinations with all other VPR methods is NetVLAD, while the least favorable technique is found to be CALC that holds low complementarity with all other state-of-the-art VPR techniques

Fig. 5 depicts the lower and upper bounds of complementarity for the different VPR combinations that are discussed above. Beginning from combinations formed with AlexNet the largest upper bounds can be observed for NetVLAD, RegionVLAD and then HybridNet. HoG has a slightly smaller upper bound value, but the smallest upper bound value is for CALC. 
AMOSNet (as the primary VPR technique), the largest upper bounds are held by NetVLAD and CoHoG, while the smallest upper bound values belong to HybridNet and CALC. The lower bounds for AMOSNet combinations are somewhat similar and remain within the 0.1 boundary. 
For combinations of CALC with other VPR techniques, the upper bounds for all combinations appear to have high values. The highest upper bound values, however, are held by NetVLAD and CoHoG and is almost 1.0. The smallest upper bound value in this case is for AlexNet and as for the lower bounds, these again appear to have similar values all below 0.2 for all combinations.    
With CoHoG (as the primary VPR technique), out of different combinations, the highest upper bound value belongs to HybridNet and AMOSNet (above 0.8). NetVLAD is also close with an upper bound value of 0.6. The lowest value is achieved by CALC which is around 0.1 quite similar to it's lower bound value. 
For HoG (as the primary VPR technique) the highest upper bound values are possessed by NetVLAD and CoHoG, while RegionVLAD and HybridNet fall next in line. The smallest upper bound value belongs to CALC and AlexNet. The lower bound score correspond to the upper bound values with the largest for NetVLAD and CoHoG and smallest for CALC and AlexNet. 
For combinations of HybridNet (as the primary VPR technique) with other state-of-the-art methods, the highest upper bound values are achieved by NetVLAD and CoHoG, while AMOSNet and RegionVLAD have significantly lower values in comparison (around 0.4 and 0.6 respectively). The lowest value, however, is still achieved by CALC and AlexNet which are almost the same as their lower bound value. 
For combinations of NetVLAD, the highest upper bounds are held by AMOSNet and HybridNet, with CoHoG next in line. As for the lower bounds, the values seem to be quite uniform and remain low for all combinations (below 0.1). 
Finally, for combinations of RegionVLAD with other state-of-the-art VPR techniques, the highest upper bound values are achieved by NetVLAD, HybridNet and AMOSNet. It may be concluded from Fig. 5 that combinations of NetVLAD appears to have the highest upper bound values for most combinations, while CALC consistently appears to have the smallest upper and lower bound values for almost all combinations that have been tested.

Table 2 presents the maximum achievable VPR performance values for 28 different combinations of state-of-the-art VPR methods utilizing the proposed framework. It is evident that each combination has varying \textit{MAPE} values over each dataset. The highest \textit{MAPE}  by a VPR combination for each dataset has been highlighted. It can be observed that for the 17Places dataset many combinations show promising \textit{MAPE}  values but there isn’t a significant difference between them, although the highest \textit{MAPE} value belongs to AlexNet + NetVLAD. For the 24-7 dataset the MAPE for almost all combinations are somewhat consistent but the highest value of 98.4\% which is by NetVLAD + RegionVLAD is significantly higher. The Corridor dataset has some varying \textit{MAPE} values ranging from 40\%-77\% while the highest value is obtained by CoHoG + HybridNet. Moving on to the CrossSeasons dataset that has lower \textit{MAPE} values overall compared to other datasets with the highest value of 36.1\% obtained by not one but three different combinations that are; AlexNet + NetVLAD, HybridNet + NetVLAD and HybridNet + RegionVLAD. The next dataset, Essex3in1, has two highest \textit{MAPE} values of 88.5\% attained by CoHoG + NetVLAD and HoG + NetVLAD. For GardenPoint dataset the \textit{MAPE} values remain between 34\% to 77\% being the highest which belongs to NetVLAD + RegionVLAD. Some of the highest \textit{MAPE} values by all combinations can be observed for the Livingroom dataset which the highest value of 100\% is obtained by AMOSNet + CoHoG and CoHoG + HybridNet both. The Nordland dataset has very diverse \textit{MAPE} values with lowest being 22.3\% by HoG + NetVLAD and the highest \textit{MAPE} value of 91.4\% by HoG + HybridNet. The SPED and SYNTHIA dataset have quite the opposite results, on one hand SPED has very high \textit{MAPE} values with the greatest value of 86.1 by AMOSNet + NetVLAD while SYNTHIA has significantly lower values for all combinations and the largest among them is 34.2 by AlexNet + NetVLAD. Overall results suggest that best options to consider after the selection of a primary VPR technique are NetVLAD, RegionVLAD and HybridNet as they appear to perform the best for most combinations they form.

\section{Conclusions and Future Work}
This paper has proposed a well-defined framework for determining the viability of combining different VPR methods for a multi-process fusion system. The complementarity information computed through the proposed framework helps to select the best possible combination of VPR techniques to ensure performance improvement in fused systems. The results obtained utilising the presented framework for eight state-of-the-art VPR methods over ten widely-used VPR datasets provide new insights regarding complementarity of various VPR methods and estimate their maximum performance. This paper has considered only pairs of VPR techniques. A promising future direction is to investigate extension to a combination of three or more VPR techniques.

\addtolength{\textheight}{-12cm}   









\end{document}